\documentclass[letterpaper]{article} 
\usepackage{aaai25}  
\usepackage{times}  
\usepackage{helvet}  
\usepackage{courier}  
\usepackage[hyphens]{url}  
\usepackage{graphicx} 
\usepackage{natbib}  
\usepackage{caption} 
\usepackage{amsfonts}
\usepackage{booktabs} 
\usepackage{algorithm}
\usepackage{algorithmic}

\usepackage{newfloat}
\usepackage{listings}
\DeclareCaptionStyle{ruled}{labelfont=normalfont,labelsep=colon,strut=off} 
\floatstyle{ruled}
\newfloat{listing}{tb}{lst}{}
\floatname{listing}{Listing}
\title{Enhancing Privacy in the Early Detection of Sexual Predators Through Federated Learning and Differential Privacy}

\author {
    Khaoula Chehbouni\textsuperscript{\rm 1,\rm 2},
    Martine De Cock\textsuperscript{\rm 3},
    Gilles Caporossi\textsuperscript{\rm 4},
    Afaf Taik \textsuperscript{\rm 2, \rm 5}\\
    Reihaneh Rabbany \textsuperscript{\rm 1, \rm 2},
    Golnoosh Farnadi \textsuperscript{\rm 1, \rm 2}
}
\affiliations {
    \textsuperscript{\rm 1}McGill University\\
    \textsuperscript{\rm 2}Mila - Quebec AI Institute\\
    \textsuperscript{\rm 3}University of Washington Tacoma\\
    \textsuperscript{\rm 4}HEC Montreal \\
    \textsuperscript{\rm 5}Université de Montréal\\
    khaoula.chehbouni@mila.quebec
}

\usepackage{bibentry}

\begin{document}

\maketitle

\begin{abstract}
The increased screen time and isolation caused by the COVID-19 pandemic have led to a significant surge in cases of online grooming, which is the use of strategies by predators to lure children into sexual exploitation. 
Previous efforts to detect grooming in industry and academia have involved accessing and monitoring private conversations through centrally-trained models or sending private conversations to a global server. In this work, we implement a privacy-preserving pipeline for the early detection of sexual predators. We leverage federated learning and differential privacy in order to create safer online spaces for children while respecting their privacy. We investigate various privacy-preserving implementations and discuss their benefits and shortcomings. Our extensive evaluation using real-world data proves that privacy and utility can coexist with only a slight reduction in utility.
\end{abstract}

%


%

\maketitle

\section{Introduction}
\label{section:introduction}

With the COVID-19 pandemic, the number of children victim of online grooming~(OG) has increased substantially: the Canadian Centre for Child Protection has recorded an 815\% spike in the reported cases of sexual exploitation online over the last 5 years~\citep{publicsafety}. The unprecedented rise in screen time and isolation brought about by the pandemic have left children more vulnerable than ever to online sexual exploitation. In 2021 alone, 85 million pictures and videos of child sexual abuse have been reported worldwide~\citep{european_commission_2022}. 

OG can be defined as the different strategies used by predators to lure children into sexual relationships. Studies have shown that predators have a particular communication style and exhibit common behavior patterns that allow them to approach children, lure them into a trusting relationship, isolate them and desensitize them to the sexual act \citep{olson2007entrapping, lorenzo2016understanding}. 
Social media platforms have changed the rules in the past decade, and made children far more accessible to predators.
The direct messages in these platforms have become a low-risk tool for luring a child into (online) sexual exploitation. 
Indeed, while parents can have a tighter hold on their children’s interactions in real life, monitoring their online conversations is far more complicated. 

In May 2022, the European Commission proposed a new regulation to compel chat apps to scan private user messages for child abuse and exploitation \citep{european_commission_2022}. This new regulation was strongly condemned by privacy experts, who believed that implementing such mechanisms and breaking end-to-end encryption of users’ messages could lead to mass surveillance \citep{vincent_2022} and compromise not only users' privacy but also their online safety: online scams, cyberattacks or social engineering attacks being more likely to happen once encryption is lifted~\citep{risktoencryption}. As a result, the e-Privacy derogation was implemented to allow companies to detect and remove child sexual abuse material online on a voluntary basis only. In April 2024, the Parliament decided to prolong the exemption until April 2026~\citep{european_commission_2024}. In the meantime, it is crucial for social media providers to devise more effective strategies for safeguarding children while also respecting the privacy of their users since previous attempts made at adopting child protection features, such as Apple's and Facebook's scanning tools, have sparked wide criticism by privacy experts~\citep{snowden_2021, facebookscan}. More recently, Google made the news when a father was the subject of an investigation for child sexual abuse and exploitation after sending naked pictures of his toddler to the  pediatrician~\citep{googleflag}. 

 Previous work in communication studies has shown that the sexual predators' discourse contains specific indicators that can be leveraged for the detection of online grooming~\citep{olson2007entrapping, lorenzo2016understanding}. Some researchers focused on finding these linguistic cues by extracting lexical, syntactical, and behavioral features from chat messages~\citep{inches2012overview, mcgheechatcoder2}. Others have used deep learning techniques to learn useful representations from text~\citep{technicalmapping,morgan2021integrating}. 
All, however, have been limited by the lack of realistic and publicly-available training datasets for the task. Although preventing grooming before any harm occurs is essential to ensure safe access to online social media platforms for children, there are only a few works that treat the grooming detection problem as an early risk detection task~\citep{lopez2018early,vogtespd}, i.e.~recognizing grooming while it is happening and intervention is possible, as opposed to detection afterward. Furthermore, most of the existing work relies on detecting online grooming by monitoring the users’ messages and none of the proposed solutions were concerned with ensuring the privacy of the training examples. This represents a major limitation for the applicability of these models in a real-life setting which is the main focus of this paper.

In this work, we explore different privacy-preserving frameworks for the early detection of sexual predators~(eSPD) in ongoing conversations. Our proposed solution for this application is the use of Federated Learning~(FL) \citep{mcmahan2017communication}, an alternative to centralized machine learning~(ML) that relies on a global server orchestrating the training of models with data originating from different entities without requiring them to share any raw data. This approach holds the potential to address the two primary challenges facing the application: obtaining access to real-world data for training while preserving privacy, and increasing the availability of labeled data through the implementation of a reporting mechanism to identify predators. However, it is important to note that while FL operates by keeping client data local to their devices during the training process, the federated architecture alone does not guarantee complete privacy protection. Studies have demonstrated that it is feasible to reconstruct a client's sensitive information from the shared updates they contribute during the process~\citep{kairouz2021advances}. We therefore enhance our solution with differential privacy~(DP) to provide formal privacy guarantees~\citep{dworkdp}.

Our work presents the following contributions: (1) different practical and privacy-preserving frameworks for eSPD; (2) a comprehensive implementation of the proposed frameworks, including a thorough examination of various DP algorithms to enhance privacy, and (3) an in-depth evaluation of the framework using a real-world dataset.


\section{Related Work}
\label{sec:relatedwork}
In this section, we review the most relevant works to our proposed approach.

\paragraph{Detection of sexual predators.} 
A competition organized at PAN-12~\footnote{https://pan.webis.de/clef12/pan12-web/index.html} attracted attention to the task of identifying sexual predators with the creation of a new annotated dataset for the detection of online grooming in messages \citep{inches2012overview}. The used techniques varied among the winners, as \citet{villatoro2012two} used neural networks and SVMs, and  \citet{popescu2012kernel} used kernel-based learning methods. 
In the following years, the problem was approached using representation learning techniques ~\citep{sentimentanalysisforspd, technicalmapping, bertworkspd, morgan2021integrating, 10.1145/3579987.3586564}.
For instance, \citet{sentimentanalysisforspd} approached the problem as a sentiment analysis task and employed recurrent neural networks with long short-term memory cells to enhance accuracy. \citet{10.1145/3579987.3586564} for their part showed how a simple contrastive learning framework for feature extraction at a sentence-level achieved state-of-the-art results. Whereas \citet{bertworkspd} used BERT to extract context-aware representations from text, similar to our approach in this work. 
However, it is worth noting that these studies approached the problem from a forensic perspective rather than a preventative one.

\paragraph{Early detection of sexual predators.} 
To block harm from occurring, grooming should be detected before a victim is lured. Escalante et al.~\cite{escalante2015early} made the first attempt at the early detection of sexual predators by adapting a naive Bayes classifier for grooming prediction with partial information. The authors evaluated the performance of their model with different percentages of words from the test set in a chunk-by-chunk evaluation framework that was later extended using profile-based representation \cite{lopez2018early}. 
More recently, \citet{vogtespd} formally defined the eSPD task, moving away from existing work to propose a sliding window evaluation, and creating a new dataset that is better suited for the task. 
The previously mentioned studies on the detection of grooming have assumed that training and deployment of models could be performed without considering privacy implications. This involves the full disclosure of users' private messages to a central server for model training. Our work extends the study by \citet{vogtespd} and utilizes their evaluation framework and dataset to demonstrate how eSPD can be adapted in a distributed setting while ensuring formal privacy guarantees.

\paragraph{Privacy-preserving text classification.}
 
Various privacy-enhancing technologies for text classification have been proposed in the literature. Solutions based on cryptographic techniques such as secure multiparty computation (MPC) or homomorphic encryption enable privacy-preserving \textit{inference}, i.e.~classification of a message held by one party with a model held by another party without requiring each of the parties to disclose their data to anyone in plain text \cite{privacyhatespeechdetection,adams2021private,smsspamdetect, lee-etal-2022-privacy}. 
However, while MPC protocols provide \textit{input privacy}, i.e.~the data holders do not disclose their inputs in plain text to anyone, the computed result -- such as the trained model -- is the same as one would obtain without the use of cryptography; in other words, MPC by itself does not provide \textit{output privacy}. This is problematic as information about the training data can leak from a model and from inferences made with it \cite{fredrikson2015model,tramer2016stealing,song2017machine,carlini2019secret}.

In this paper, we use a combination of FL and DP to provide both input \textit{and} output privacy. While the use of FL and DP has received a considerable amount of attention in the natural language processing (NLP) literature (see e.g.~\cite{whatdoesitmean,klymenko2022differential} and references therein), only a few authors have looked into the use of FL for supervised learning tasks of text classification. 
In particular, FL an DP were combined in applications such as sentence intent classification \citet{zhu2020empirical} and financial text classification \citet{basu2021privacy}. More recently, \citet{Shetty_Muniyal_Priyanshu_Das_2023} and \citet{samee2023safeguarding} leveraged FL for privacy-preserving detection of cyberbullying: \citep{Shetty_Muniyal_Priyanshu_Das_2023} introduced FedBully, a framework leveraging FL, sentence-encoders and secure aggregation for privacy-preserving detection of cyber-bullying whereas \citep{samee2023safeguarding} looked at how FL with DP, word embeddings and emotional features could be paired for cyberbullying detection. 
We focus instead on evaluating how different DP implementations can affect a framework for the early detection of abusive content, and the challenges of such implementations in a real-life scenario.

\section{Background}
\label{sec:background}
In our work, we leverage FL and DP to protect the privacy of users. Hence, we first introduce FL and then provide a brief overview of the different DP algorithms we implement. 


\paragraph{Federated learning.} Introduced by \citet{mcmahan2017communication} as an alternative to privacy-invasive centralized learning, FL is an ML technique that allows multiple entities, called clients, to collaboratively learn a statistical model under the coordination of a central server. The global server orchestrates the training by sampling, at each round, a set of clients to participate in the training. Each selected client downloads the current global model, trains it further on its local data and shares a focused update with the server. The server then collects and aggregates all the updates before updating the global model. The aggregation algorithm used by the global server plays an important role in the federated setting since it defines how the training is orchestrated and how the final model will be computed. In this work, we use the federated averaging algorithm~(FedAvg)~\citep{mcmahan2017communication}.



\paragraph{Differential privacy} Whilst FL protects the privacy of the clients by preventing the sharing of raw data, FL in itself does not offer formal privacy guarantees, and the resulting model can leak information about the training data~\citep{thakkar2021understanding,memorization}.
To mitigate such information leakage, FL can be combined with DP~\citep{dworkdp} to provide plausible deniability regarding the existence of an instance or a user in a dataset, i.e.~offering protection against membership inference attacks~\citep{membershipinferenceattacks}. 

DP revolves around the idea of a randomized algorithm -- such as an algorithm to train ML models -- producing very similar outputs for adjacent inputs. Formally, a randomized algorithm 
$\mathcal{M}: D \mapsto R$ with domain $D$ and range $R$ is said to be  ($\epsilon$, $\delta$)-differentially private if for any adjacent datasets $x$ and $x'$ and for all subsets of outputs $S \subseteq R$ we have $Pr[\mathcal{M}(x) \in S] \leq e^\epsilon  Pr[\mathcal{M}(x') \in S] + \delta$, where $\epsilon$ measures the privacy loss (privacy budget) whereas $\delta$ is the probability of data being accidentally leaked. 
The smaller these values, the stronger the privacy guarantees. 

DP can be implemented at the \textit{instance-level}: two datasets are considered adjacent if they only differ in one labeled instance, or at the \textit{user-level}: two datasets are considered adjacent if they only differ in one user's complete data. 
According to the above definition, $\mathcal{M}$ is then ($\epsilon$, $\delta$)-DP if the probability that $\mathcal{M}$ generates a specific model from the data is very similar to the probability of generating that model if a particular instance (resp.~user) had been left out of the data. The latter implies that what the model has memorized about individual instances (resp.~users) is negligible.


DP is a widely used privacy-preserving technique in the field of machine learning. DP provides formal privacy guarantees for sensitive data by adding controlled noise to the data, thereby obscuring individual contributions.There are two main types of DP: local DP and global DP. Local DP operates on individual data points, whereas global DP operates on aggregate data. And whereas global DP relies on the notion of adjacent inputs, local DP does not. As such, a randomized algorithm $\mathcal{M}$ is said to satisfy $\epsilon$-Local DP if and only if for any pair of input values $v$ and $v'$ in the domain of $\mathcal{M}$ and for any output $y \in \mathcal{Y}$ we have $ \mathbb{P}[\mathcal{M}(v) = y] \leq e^\epsilon  \cdot \mathbb{P}[\mathcal{M}(v') = y]$, where $\mathbb{P}[\cdot]$ is the probability.

DP can be applied in various stages of the machine learning pipeline: either during pre-training by adding noise to the raw data, or during training by injecting noise into gradients or weights, as further outlined below.



\textbf{\textit{Metric DP}}
Local differential privacy~(LDP) allows each user to perturb their own data, providing strong privacy guarantees, typically at a cost in utility. Indeed, in the context of language, applying LDP means that all word embeddings are exchangeable without accounting for meaning, which may hurt the utility of the downstream task. 
To tackle this challenge, \citet{feyisetandxprivacy} adapt a relaxed version of DP, called metric DP, for natural language processing. A randomized mechanism $\mathcal{M} : \mathcal{D} \mapsto \mathcal{R}$ is said to be $\eta d_{\chi}$-private if for any $x$ and $x' \in \mathcal{D}$ the distribution over outputs of 
$\mathcal{M}(x)$ and $\mathcal{M}(x')$ is bounded by: $\mathbb{P}[\mathcal{M}(x) = y] \leq e^{\eta d(x,x')}  \cdot \mathbb{P}[\mathcal{M}(x') = y]$
for all $y \in \mathcal{R}$, where $d$ is a distance function $d: \mathcal{D} \times \mathcal{D} \mapsto \mathbb{R}_+$. 
\citet{privacypreservingbert} adopt this definition for sequence representation privatization: the input to the randomized mechanism $\mathcal{M}$ is a token embedding, or in our case, the [CLS] representation produced by BERT. As such, metric DP can be achieved for the Euclidean distance by adding calibrated noise
to the sequence representations of the training data~\citep{privacypreservingbert}. 

\textbf{\textit{The DP-SGD Algorithm}} An $(\epsilon,\delta)$-DP randomized algorithm $\mathcal{M}$ is commonly created out of an algorithm $\mathcal{M^*}$ by adding noise that is proportional to the sensitivity of $\mathcal{M^*}$, in which the sensitivity measures
the maximum impact a change in the underlying dataset from $d$ to an adjacent dataset $d'$ can have on the output of $\mathcal{M^*}$. 
DP-SGD is a commonly used technique for training deep neural networks with DP guarantees. DP-SGD is designed to limit the impact that the training data has on the final model by making the mini-batch stochastic optimization process differentially private through gradient clipping and adding noise~\citep{Abadi_2016}. 


\textbf{\textit{The DP-FedAvg Algorithm}} \citet{mcmahan2017} introduce a DP version of the FedAvg algorithm to offer user-level privacy.
Since the FedAvg algorithm already groups multiple SGD updates together and operates at the user-level, it is possible to extend the DP-SGD algorithm to provide formal privacy guarantees to federated training. 
In order to achieve ($\epsilon, \delta$)-DP, each client clips its own update, and after each round of federated training, the server adds noise to the aggregated updates. The overall privacy cost relies on the number of clients sampled at each round of training, on the number of federated rounds of training, and on a hyperparameter that controls the scale of the noise added to the update: the \textit{noise multiplier}.

\section{Method}
\label{sec:method}
Ensuring the safety of children from cybercrime is crucial, but it should not compromise the privacy of social media users. As such, we introduce a privacy-preserving framework for the identification of sexual predators. 
This novel approach takes advantage of the rising prevalence of mobile devices among children and teenagers.

\subsection{eSPD Training via Federated Learning}
\label{subsec:espdviafederatedlearning}

One of the significant challenges in FL is addressing non independent and identically distributed (IID) data, as the local data distribution of each client is not representative of the overall population~\citep{zhu2021federated}. This challenge becomes even more pronounced in the context of OG, where most users are unlikely to interact with sexual predators. Thus, the detection of OG in a federated setting can be viewed as an extreme case of non-IID data where most users will only have access to one label for training. 
To address this issue, we implement a data-sharing strategy during training in which a small portion of \textit{warm-up data} is distributed to each device in addition to the initial model~\citep{zhao2018federated}. The \textit{warm-up data}, which contains public examples from both classes and is balanced, can be seen as a starting point for training and helps alleviate the statistical challenge.


\subsection{eSPD Training with Differential Privacy}
\label{subsec:privacypreservingframework}

In this work, we explore how FL and DP can be used to offer formal privacy guarantees in different ways. Figure~\ref{fig:privacyimplementations} illustrates our different privacy-preserving frameworks. 

\begin{figure}[t]
\centering
\includegraphics[width=1\columnwidth, height =6cm]{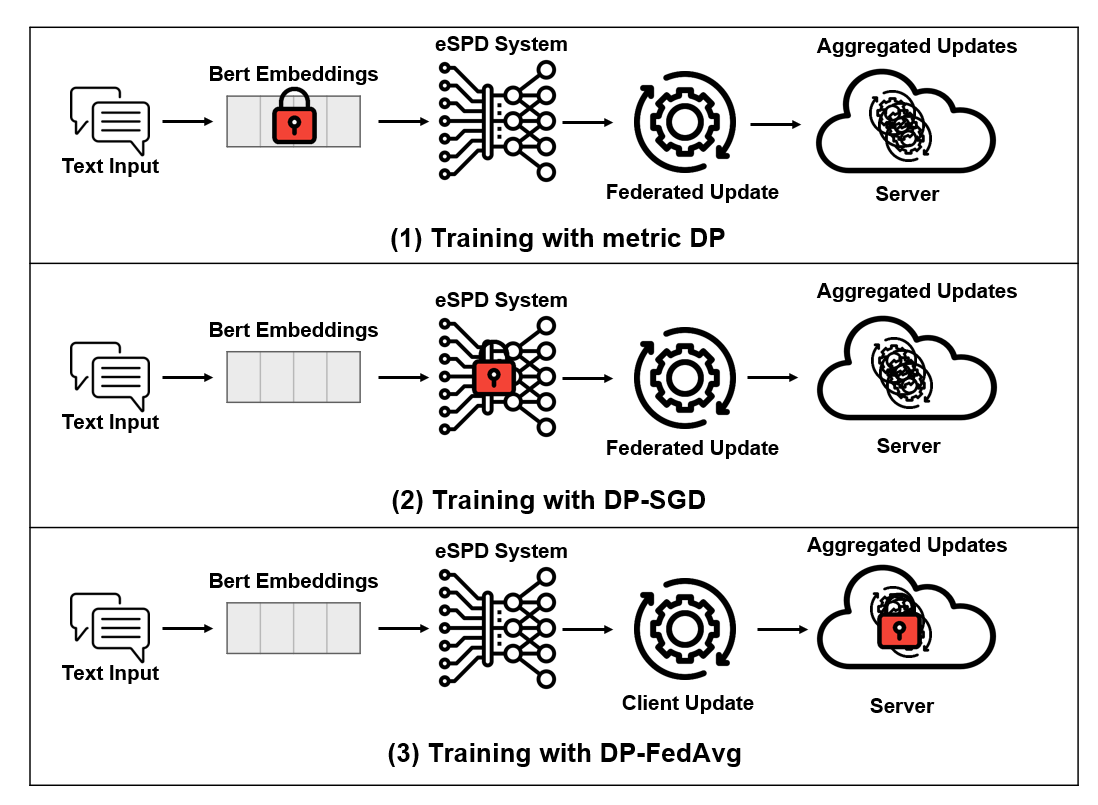} 
\caption{Illustration of the three DP implementations described in this work. Each step of the training process is represented: the learned BERT representation, the local training, and the global training. A lock representing DP is applied first to the embedding representation, then to the local training, and finally to the global training.}.
\label{fig:privacyimplementations}
\end{figure}

\paragraph{(1) Adding noise to the raw data.} 
Each client adds noise directly to the token representations of their data to achieve sequence representation privatization. The perturbed embeddings are then used for training, thus ensuring no access to the raw data. In the rest of this paper, we refer to this implementation as \textit{FL with metric DP}.

\paragraph{(2) Adding noise during training.} 
Each client train their model using DP-SGD to mitigate leakage of personal information to the server. At each round of local training, the clients clip the gradient norm of outliers in their data and adds calibrated noise to the gradients to ensure that the shared updates provide instance-level privacy protections. In this case, the raw data is accessed during local model training, but never sent to a global server. In the rest of this paper, we refer to this implementation as \textit{FL with DP-SGD}.

\paragraph{(3) Adding noise to the updates.} We train our federated model using the DP-FedAvg algorithm. After each round of training, the server adds noise to the aggregated updates sent by the clients. This approach provides user-level protections but relies on the hypothesis that the server can be trusted since it has access to non-private updates. In the rest of this paper, we refer to this implementation as \textit{FL with DP-FedAvg}.

\section{Evaluation}
\label{sec:evaluation} 
In this section, we present an empirical evaluation of our different frameworks for the early detection of sexual predators. All our experiments were performed on the PANC dataset~\citep{vogtespd}. 



\paragraph{Data.} A main challenge in addressing the sexual predators’ identification task through ML comes from the lack of publicly available labelled and realistic datasets. Only two distinct datasets are available: the PAN12 dataset~\citep{inches2012overview} and the ChatCoder2 dataset~\citep{mcgheechatcoder2}, and both take their grooming examples from the Perverted Justice~(PJ) website\footnote{http://www.perverted-justice.com/}. 
The PAN12 dataset was created by sampling non-grooming examples from IRC logs and the Omegle forum\footnote{ https://www.omegle.com} and grooming examples from the PJ website. Every conversation was then split into segments of 150 messages or less assigned to a user. The ChatCoder2 dataset, on the other hand, only contains grooming examples: 497 complete conversations also extracted from PJ.   
Since the eSPD task requires both complete conversations as well as grooming and non-grooming examples, \citet{vogtespd} introduced the PANC dataset as a better alternative for the task. They merged the non-grooming examples from the PAN12 dataset with the grooming examples from ChatCoder2. To standardize the dataset, they also split every grooming conversation into multiple segments of 150 messages or less assigned to the same user. Table~\ref{tab:tabpancover} in Appendix~\ref{sec:datastats} presents statistics about the dataset.


\paragraph{Inference of an eSPD System.} The considered inference phase is similar to \citet{vogtespd}'s framework for eSPD.
We use a sliding window for the sequential classification of a conversation. Here, a conversation consists of a sequence of messages $t_1,t_2,\ldots$. For a window of length $l$, at step $s$ the classifier labels the sequence $t_s,t_{s+1},\ldots,t_{l-1}$, at step $s+1$ the classifier labels the sequence $t_{s+1},t_{s+2},\ldots,t_l$. 

After every window prediction, the system decides whether to raise a warning or not based on the inferred labels of the last 10 window predictions. If a pre-defined threshold -- called skepticism level -- is reached, a warning is raised and the whole conversation is classified as a grooming conversation. An eSPD system never classifies a conversation as non-grooming if there are messages left, or if it is still ongoing. 
Figure~\ref{fig:inference} in Appendix~\ref{sec:illustration} illustrates this framework.

\paragraph{Evaluation Metrics.} We use the latency-weighted F1 score~\citep{sadeque2018measuring, vogtespd} to evaluate our models. The F-latency score measures the trade-off between the speed of detection (i.e.~how early in a conversation grooming is detected) and the accuracy of the warning by applying a penalty that increases with the warning latency. The warning latency is defined as the number of messages exchanged before a warning is raised~\citep{vogtespd}. As such, \textit{a higher F-latency score means a better-performing eSPD system}. The penalty can be computed for each warning latency $l \geq 1$ as follows: 
\begin{displaymath}
\textrm{penalty}(l) = -1 + \frac{2}{1+e^{(-p \cdot (l-1))}}
\end{displaymath}
where $p$ defines how quickly the penalty should increase. As suggested by \citet{sadeque2018measuring}, $p$ should be set such that the latency penalty is $50\%$ at the median number of messages of a user. 

The $speed$ of an eSPD system over a test set of grooming conversations is defined as $\textrm{speed} = 1 - \textrm{median}\{\textrm{penalty}(l)  \ | \ l \in \textrm{latencies}\} $ where $latencies$ corresponds to the list of warning latencies produced by the system for all grooming conversations for which a warning is raised. We can then formally define F-latency as: $\textrm{F-latency} = \textrm{F1} \cdot \textrm{speed}$.

While F1 is computed across the entire test set of positive and negative messages, penalty and speed are computed for the positive conversations only. This is common practice in the literature
as the delay needed to detect true positives is a key component of the early risk detection task \citep{losada2020overview, sadeque2018measuring}.


\paragraph{Choice of the classifier.} Although fine-tuning BERT~\citep{devlin2019bert} has been shown to give better results for eSPD~\citep{vogtespd} (see Appendix~\ref{sec:comparisonlit}), we use the pre-trained feature-based approach with logistic regression~(LR) for our experiments since it is far less computationally expensive and better suited for scaling federated training to a large number of clients. 
In our framework, each user uses the $\textrm{BERT}_\textrm{BASE}$ model to create a context-aware representation of their personal conversation by extracting fixed features from the pre-trained model. The [CLS] representation of the last layer is then used as an input for LR with a binary cross entropy loss function. For each user's segment, we obtain a $768$ length vector.

\section{Empirical Results and Discussion} 
\label{sec:results}

In this section, we investigate three research questions. Details about our implementation can be found in Appendix~\ref{sec:experimentalsetup}, and Appendix~\ref{sec:addresults} presents additional quantitative results. Our eSPD implementation is available online \url{https://github.com/khaoulachehbouni/fl-espd}. 

\subsection{RQ1: How is the utility of the eSPD system affected by the FL framework?} 

We compare the utility of our \textit{FL model} with two baselines: 
(1) \textit{Warm-up  model}: A BERT+LR model trained on the warm-up data only, to ensure that the federated frameworks are not too biased by the data-sharing strategy; and (2) \textit{Centralized model}: A BERT+LR model trained centrally on the training data and the warm-up data.



In Table~\ref{tab:espdresults}, we see that the federated model shows competitive results for the eSPD task with the highest F1 score (82\%) and F-latency score (64\%). 
The FL model also have the lowest false positive rate~(FPR) with 3\% showing that the FL framework is indeed suited for the accurate detection of predators. Finally, despite having a very high speed, the baseline \textit{Warm-up model} has the lowest F-latency score: our model is therefore not biased by the data-sharing strategy and it is indeed learning from each client’s personal data.

\begin{table}[t]
\centering
\resizebox{\columnwidth}{!}{%
\begin{tabular}{lcccccc}
\hline
\textbf{Model} & \textbf{F1} & \textbf{Recall} & \textbf{Precision} & \textbf{Speed} & \textbf{F-latency} & \textbf{FPR} \\
\hline
Warm-up model & 0.50 & \textbf{0.98} & 0.33 & \textbf{0.96} & 0.48 & 0.24 \\
Centralized model & 0.75 & 0.95 & 0.62 & 0.83 & 0.63 & 0.07 \\ 
FL model & \textbf{0.82} & 0.85 &\textbf{ 0.79} & 0.79 & \textbf{0.64} & \textbf{0.03} \\ 
\hline
\multicolumn{7}{c}{\textbf{Evaluation results for a 1\% FPR}}\\
\hline
Warm-up model & - & - & - & - & - & - \\

Centralized model & \textbf{0.85} & \textbf{0.83} & 0.88 & 0.69 & 0.59 & 0.01\\

FL model  & 0.83 & 0.78 & \textbf{0.89} & \textbf{0.73} & \textbf{0.61} & 0.01 \\ 

\hline
\end{tabular}}
\caption{Evaluation results for the eSPD task}
\label{tab:espdresults}

\end{table}

\subsection{RQ2: How to reduce the harm of false positives in eSPD?} 
In eSPD, the emphasis is often put on the detection of predators since missing one could cause a lot of harm. Indeed, the F-latency score depends on both the F1-score and the speed. And while the F1-score takes into consideration both recall and precision, detection speed comes at the cost of precision as shown in Table~\ref{tab:espdresults}. In a real-life setting, we expect an alarm to be raised each time a predator is detected.  Given the nature of grooming messages, it is more likely that an eSPD system will be mistaken when confronted with sexual or intimate conversations, which could lead to innocent people being wrongly accused (e.g. sex workers).

Furthermore, in a real-life scenario, data often originates from forensic evidence after a predator's conviction. Given the over-representation of certain demographic groups in prison because of the failures of the criminal justice system~\citep{govcan_black, govcan_indigeneous}, we could imagine a model making spurious correlations between vernacular and predatory behavior, leading to biases against specific demographic groups.
Therefore, we propose an approach to consider the cost of falsely accusing someone as a predator. For this purpose, for each of our models, we identify the classification threshold needed to achieve a $1\%$ FPR when evaluated on the test set. Using this new threshold, we re-evaluate our models. Table~\ref{tab:espdresults} shows that varying the threshold comes with a loss in speed, which is to be expected since higher prediction scores are now needed to classify a window as OG. Furthermore, the results for the baseline \textit{Warm-up model} are not presented because the smaller FPR attained for this model with a 0.99 classification threshold is 9\%: the model is therefore falsely classifying non-OG conversations as OG. Finally, we notice a decrease in F-latency for all the models, a necessary trade-off to achieve better precision.

\subsection{RQ3: What are the benefits and shortcomings of different privacy implementations for eSPD?}

\begin{table*}[t]
\caption{Evaluation results for the eSPD task with different privacy implementations}
\label{tab:espdresultsdp}
\centering
\resizebox{\linewidth}{!}{
\begin{tabular}{l | cccc| cccccc}
\hline
& \multicolumn{4}{c|}{\textbf{Privacy Protection}} & \multicolumn{6}{c}{\textbf{Utility}}\\
\textbf{Model} & \textbf{Input} & \textbf{Output} & \textbf{Type} & \textbf{Level} & \textbf{F1} & \textbf{Recall} & \textbf{Precision} & \textbf{Speed} & \textbf{F-latency} & \textbf{FPR}\\
\hline
FL with metric DP  & Input & Output & Local & Instance& 0.73 & 0.64 & 0.84 & 0.71 & 0.52 & 0.02 \\
FL with DP-SGD & Input & Output & Local & Instance & 0.77 & 0.86 & 0.70 & 0.79 & 0.61 & 0.04 \\
FL with DP-FedAvg   & Input & Output & Global & User & 0.81 & 0.86 & 0.76 & 0.78 & 0.63 & 0.03 \\\hline
\end{tabular}}
\end{table*}

Language data, especially personal conversations on social media, is very sensitive for two main reasons. First, it contains a lot of personally identifiable information (PII): people share their name, number, address, and personal details about themselves through text messaging. Second, studies have shown that beyond PII, it was possible to recover information about the author of a text based on linguistic patterns~\citep{mattern-etal-2022-limits}. For these reasons, and given the sensitivity of the eSPD application, we look at how different privacy implementations impact our eSPD system and what it means in a real-life scenario. 
In this work, we have discussed four different ways to offer privacy protections in the context of eSPD, and each comes with its benefits and shortcomings.

\paragraph{\textbf{The FL implementation.}}

Even though this implementation does not guarantee privacy, the federated framework offers additional protection compared to the normal centralized setting since each user's raw data never leaves their device. However, information about the training data can still be recovered from the trained model. Given the sensitive nature of personal conversations on social media, a federated framework alone may not be enough for eSPD.

\paragraph{\textbf{The FL with metric DP implementation.}}
We first consider a scenario where each user adds noise to their private data. As such, even during local training, the service provider never has access to the raw data. Table~\ref{tab:espdresultsdp} shows that the \textit{FL model with metric DP} has a 52\% F-latency with an $\eta=20$. Even with our oversampling technique during training, injecting noise directly into the training data comes with a high utility cost. In Figure~\ref{fig:figureldpprivacy} in Appendix~\ref{sec:experimentalsetup}, we can see that models trained with more restrictive privacy budget (e.g. $\eta=10$) are non-performing (e.g. with an F-latency score of 6\%).

This implementation however offers strong privacy guarantees since it gives no access to the raw data even for local training. However, metric DP might not be the most appropriate implementation in the case of eSPD. Indeed, bounding the privacy budget to a distance measure still gives information about the neighborhood of the original input: a sexual message for example is still a sexual message if not enough noise is added to the data. Simultaneously, having a more restrictive privacy budget have a high cost on the utility of the eSPD system since it means increasing the semantic distance between the original input and the perturbed one.



\paragraph{\textbf{The FL with DP-SGD implementation.}} To add formal privacy guarantees to the FL framework, we implement DP-SGD. DP-SGD offers strong privacy protection against membership inference attacks by perturbing the training of a model. While in this scenario, the service provider still has access to the raw data, the data never leaves a user's device, and the updates aggregated for federated training are private. By training privately on each user's device, we guarantee instance-level DP. In our case, the instances being each data point shared by the users. 

This implementation also comes with a cost in utility. We experiment with different epsilon bounds at the users' levels and we see that with ($\epsilon=0.50, \delta=10^-5$) we register a drop in utility with a 57\% F-latency score. In Table~\ref{tab:espdresultsdp}, we can see that with ($\epsilon=1, \delta=10^-5$) we can still achieve a f-latency score of 61\%. It is important to note however that the privacy budget computed is not the total privacy budget for the implementation, since the same client can be resampled.

In text classification with FL, each user is expected to share thousands of data points, and instance-level DP only offers plausible deniability that each data point was not part of the training rather than protecting the user in itself. This might not be the most appropriate implementation for eSPD, since we expect a child's entire conversation history to be protected, instead of individual messages.

\paragraph{\textbf{The FL with DP-FedAvg Implementation}} 
In DP-FedAvg, each user train a model locally and send non-private updates to the server. The server aggregates these updates and makes them private by adding noise. This implementation rely on the assumption than the server is trustworthy and that the updates are not intercepted. Training with DP-FedAvg usually has less of an impact on the utility of the model. Indeed, by adding noise after the training is done, the DP-FedAvg algorithm ensures privacy guarantees with almost no utility loss as we can see in Table~\ref{tab:espdresultsdp} with an F-latency score of 62\% for a total privacy budget of ($\epsilon=1, \delta=10^-5$). Indeed, it has been shown that given sufficient local computation on each client's data, DP-FedAvg has a negligible impact on utility~\citep{mcmahan2017}, as such, we can achieve similar utility by sampling fewer clients at each round. 

Finally, training with DP-FedAvg ensures user-level protection since it offers privacy guarantees for all the data points shared by one user, which is essential in the context of eSPD. Indeed, since each user may contribute thousands of data points, having instance-level protections may not be enough. However, this approach does not protect the clients from the server since non-private updates are exchanged. Furthermore, as DP guarantees rely on amplification-via-sampling, a large number of clients is needed for a DP-FedAvg model to achieve good utility, which may be problematic in our case since labeled data is scarce in eSPD.

\section{Conclusion and Future Directions}
\label{sec:conclusion}

In December 2023, Meta announced the implementation of end-to-end encryption for one-to-one messages and voice calls on Messenger and Facebook~\citep{metaencryptionchild}, a response to public backlash following its cooperation in a Nebraska abortion case where personal chat data was turned over to the police~\citep{facebookabortion}. This decision has faced opposition from child safety groups, who view end-to-end encryption as a setback to detecting online grooming on social media~\citep{metaencryptionchild}. However, EU lawmakers do not consider end-to-end encryption and child protection as conflicting, evident in their regulations compelling major tech companies to identify and remove online child abuse material~\citep{eulawmakers}. Finding alternatives to current privacy-invasive monitoring systems is therefore critical to ensure children's safety.

In this work, we propose different federated learning frameworks for the early detection of sexual predators as a solution for both the lack of available labelled datasets and the privacy challenges that come with such an application. Indeed, by combining different implementations of differential privacy in our framework, we show that children's safety does not necessarily mean sacrificing privacy. 
We aim for our work to highlight the significance of this application and encourage the creation of more datasets and research in this area. Future research could also examine the integration of various privacy-enhancing technologies to secure against the possibility of encryption backdoors or client-side scanning. This can be achieved through techniques such as secure aggregation~\citep{secureaggregationflower} or homomorphic encryption~\citep{homomorphicencryptionfl}. Furthermore, we believe that our framework has the potential for broader application, such as in the early detection of cyberbullying or depression.

\section{Limitations and ethical considerations}
\label{sec:limitationsandethics}
One of the main challenges of the eSPD task comes from the lack of publicly available labeled and realistic datasets. The different datasets used in the literature all take their grooming examples from the PJ website, which are examples of conversations between predators and adults posing as children to catch them. Such chats have been shown to differ from real-life conversations and lack certain aspects of grooming like overt persuasion and sexual extortion \citep{schneevogt2018perverted}. Indeed, volunteers are often actively trying to get the offenders to be sexually explicit and to arrange an encounter, which is not the case in real-life settings. Furthermore, the non-grooming examples often come from forums and chatrooms where strangers can interact or engage in cyber sex, as opposed to conversations among family members, friends or partners.

We hope that the frameworks we propose in this paper will give access to a larger range of training examples. Indeed, since each user will be given the option to report abusive content, the conversations flagged as alleged grooming will then be added to the pool of training examples, thus alleviating the lack of realistic and available labelled datasets. Such a system will allow the training examples to be updated regularly and will consider the growing speed at which language, especially internet slang, evolves. We believe such challenges have hindered progress in our community’s efforts in this impactful domain. 

However, we can imagine that even with such a framework, the labelling will still be an issue since it will rely on users' self-reporting cases of grooming. We could think of a preliminary training phase with real data of convicted predators before deploying a pre-trained model to evaluate each user’s personal conversation and send a notification where a warning is raised by the eSPD system. This model will also alleviate the privacy cost since the first training phase will happen on publicly available data. In this setting, the user will be able to give feedback on the model’s prediction. But this setup is certainly not ideal, since actual victims of online grooming often trust their abuser and may not realize that they are being manipulated. Notifying a third party, either a legal guardian or a social worker tasked with monitoring the flagged content may increase the chances of a case of grooming being reported but will undoubtedly infringe on the privacy of the victim. 

Furthermore, it is important to understand how privacy guarantees in a machine learning context are not equivalent to privacy as a social norm~\citep{whatdoesitmean}. Indeed, the data protection brought by DP considers that the
personal information that needs to be protected has clear borders. In our case, for example, it means that we assume that a user is protected if all its data is. However, in a real-life scenario and particularly when it comes to language, personal information about a user can be contained in another user's data. When deploying such a system, it is important to not make false promises to potential users, and explain that protecting their data does not necessarily mean protecting their privacy, as they may understand it in a non-ML context.

Involving law enforcement could also have disastrous consequences. The resulting model could be biased towards certain populations like sex workers, people from the LGBTQI+ community, or people prone to online dating. Evaluating and selecting the best model based on a classification threshold that guarantees a $1\%$ false positive rate can be a first step towards ensuring that the eSPD system does not falsely incriminate. 
Furthermore, language models have been shown to reproduce racial and gender biases \citep{liang2021towards}. As such, using such models as a basis for identifying potential suspects to be prosecuted could lead to unanticipated outcomes. Such a system should therefore never be used directly by law enforcement agencies at the risk of
exacerbating existing social inequalities and persecuting innocents.

\appendix

\section{Statistics About the Dataset}
\label{sec:datastats}

The PANC dataset (see Table \ref{tab:tabpancover}) was split into a training set (60\%) and a test set (40\%). The training set consists of 1,753 positive segments (representing in total 298 full-length positive conversations and 9\% of the training examples) and 17,598 negative segments, whereas the test set contains 11\% examples of grooming. 

\begin{table*}
\centering
\begin{tabular}{c |rr |ll|l rr}
\toprule
      & \multicolumn{2}{l}{Number of segments} & \multicolumn{2}{l}{\# Words/segment}   & \multicolumn{2}{l}{\# Messages/segment}      \\
\midrule
Label & train           & test                   & train            & test       & train            & test                           \\
\midrule
 0     & $17,598$ $(91\%)$ & $11,733$ $(89\%)$ & $173$ $(\pm{1,385})$ & $184$ $(\pm{1529})$ & $36$ $(\pm{25})$  & $36$ $(\pm{26})$\\
1     & $1,753$ $(9\%)$   & $1,426$ $(11\%)$    & $289$ $(\pm{218})$  & $292$ $(\pm{222})$   & $64$ $(\pm{43})$  & $65$ $(\pm{43})$\\
\bottomrule
\end{tabular}
\caption{Statistics about the PANC dataset~\citep{vogtespd}.}
\label{tab:tabpancover}
\end{table*}

\section{Illustrations of our framework}
\label{sec:illustration}

\subsection{The Training Phase}
\label{subsec:trainingphase}

\begin{figure}[H]
    \centering
    \includegraphics[width=1\linewidth]{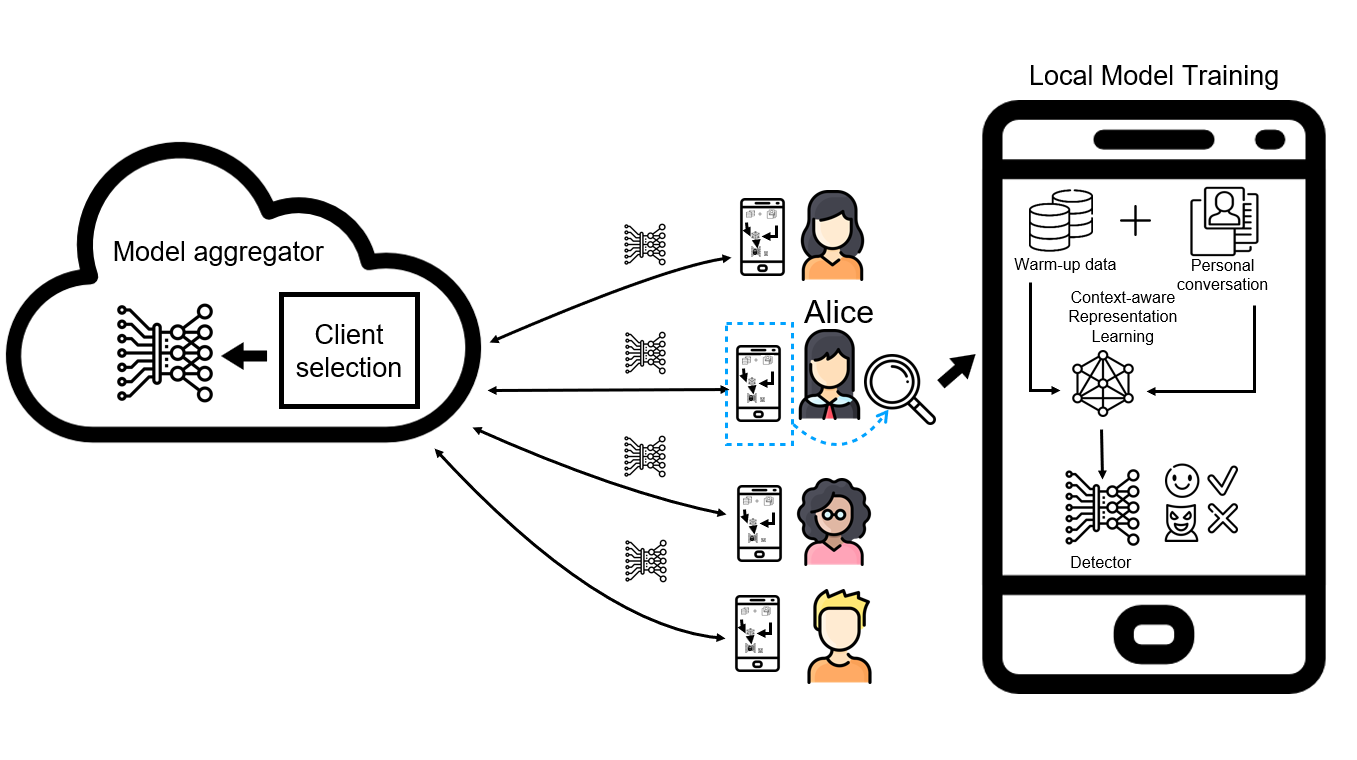}
    \caption{eSPD: Training Phase}
     \label{fig:training}
\end{figure}

Figure~\ref{fig:training} illustrates the training phase of our framework. A global server selects clients to participate and distributes a model to them; the clients will then further train the model in a privacy-preserving manner on their mobile devices using their own personal data as well as a portion of warm-up data, as we can see in Alice's cellular device.

\subsection{The Inference Phase}
\label{subsec:infphase}

\begin{figure}[H]
    \centering
    \includegraphics[width=1\linewidth]{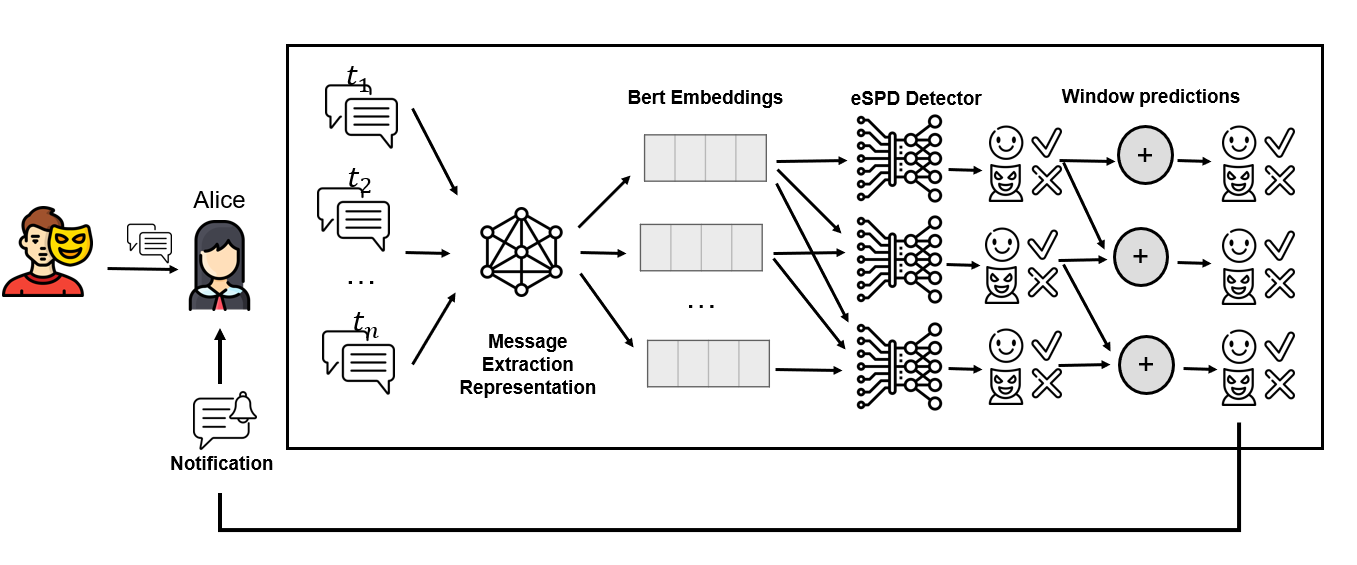}
    \caption{eSPD: Inference Phase}
    \label{fig:inference}
\end{figure}

In Figure~\ref{fig:inference}, we show how different messages received by Alice are analyzed by first being turned into word embeddings and then passed to a classifier given a sliding window for classification. Note that the final prediction is determined based on the previous sequence of predictions and that a warning notification is triggered only when multiple messages are sequentially classified as being grooming messages.

\subsection{The Privacy Implementations}
\label{subsec:privacy}

\begin{figure}[t]
\centering
\includegraphics[width=1\columnwidth]{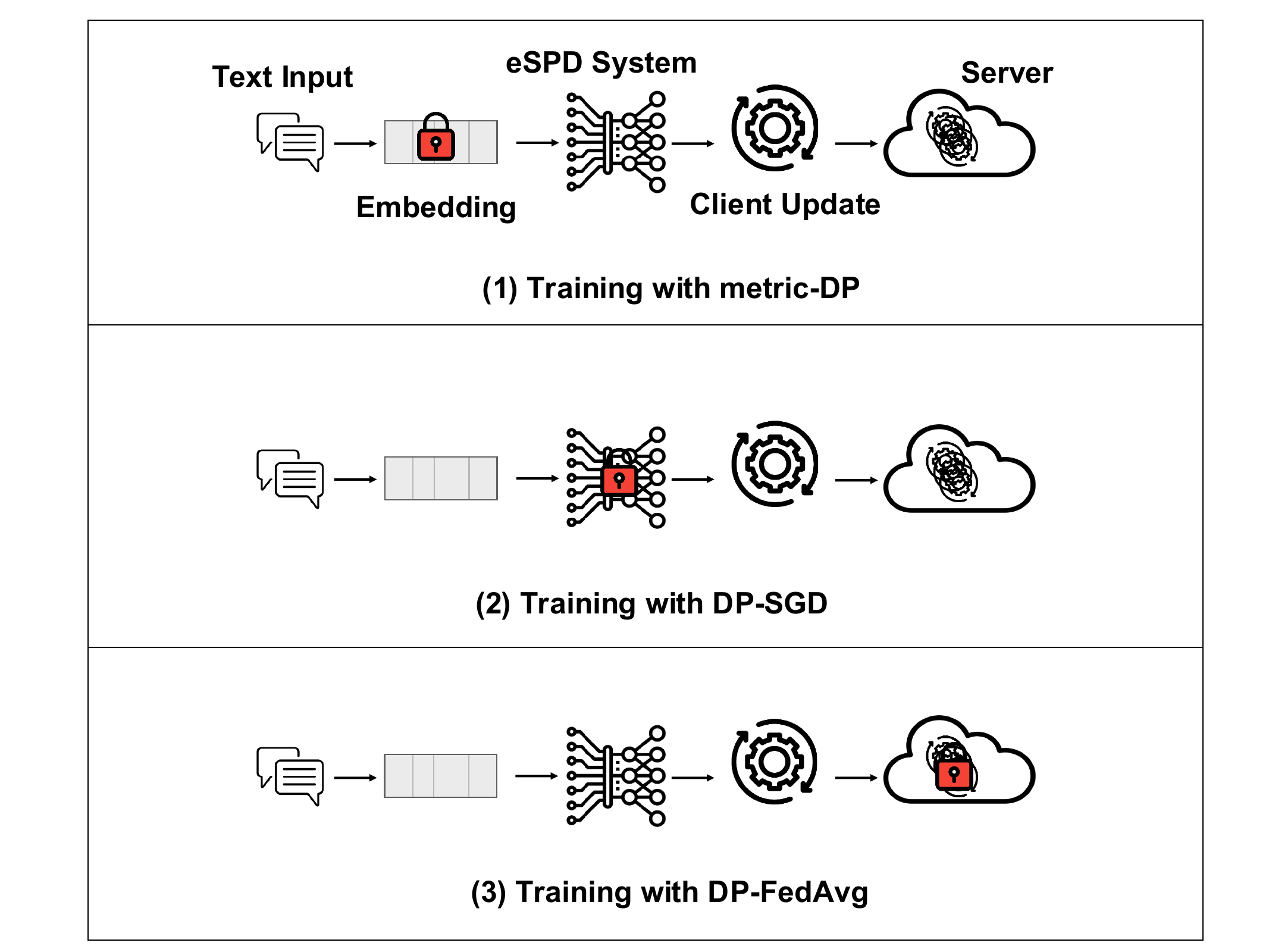} 
\caption{Illustration of the different privacy implementations: a lock representing DP is applied first to the embedding representation, then to the local training, and finally to the global training.}.
\label{fig:privacyimplementations}
\end{figure}

\subsection{The Deployed System}
\label{subsec:deployed}

\begin{figure}[t]
\centering
\includegraphics[width=0.6\columnwidth]{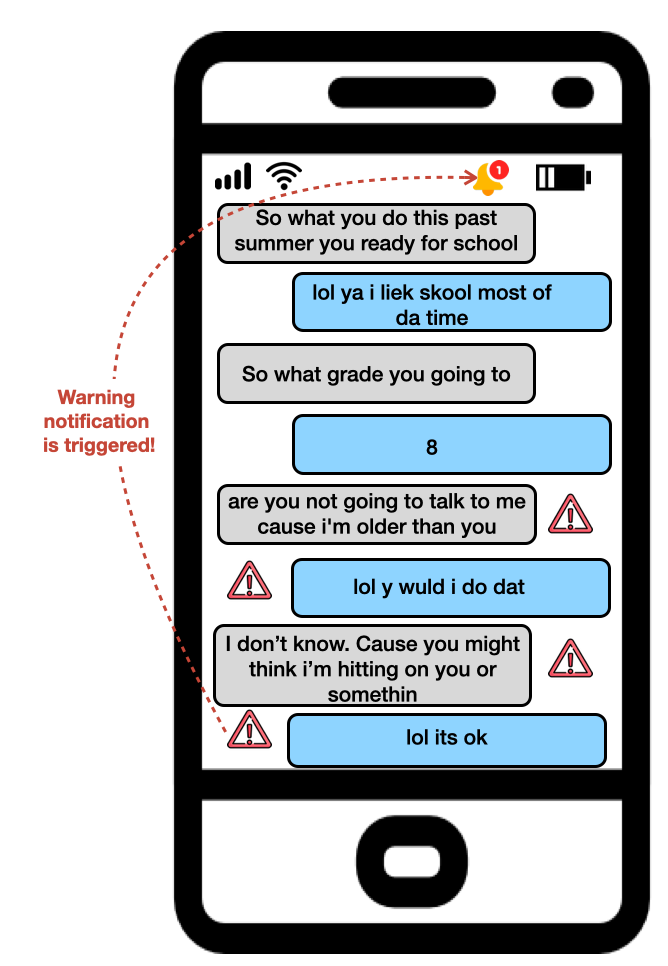} 
\caption{Visualization of eSPD in which the risk is detected, a warning is raised after passing a threshold, and the user is notified as early as possible.}.
\label{fig:eSDPvisualization}
\end{figure}

In Figure~\ref{fig:eSDPvisualization}, we present a visualization of a synthetic setup based on the proposed framework using a predatory conversation from the PANC dataset. It can take weeks or even months before a warning notification is triggered when a child is being lured by an abuser. Our goal is to minimize the harm by detecting the abuse early and sending a notification to the user. It is up to the user to decide whether to continue the conversation or report the predator. Note that in our framework, both training and inference phases are happening locally and users' personal conversations are never shared with a third party. Moreover, the global aggregated model from the server can further be tuned and personalized based on users' local data.
In Figure~\ref{fig:inference}, we show how different messages received by Alice are analyzed by first being turned into word embeddings and then passed to a classifier given a sliding window for classification. Note that the final prediction is determined based on the previous sequence of predictions and that a warning notification is triggered only when multiple messages are sequentially classified as being grooming messages.

\section{Experimental Set-Up}
\label{sec:experimentalsetup}

In this section, we present the experimental details of our implementation.


\paragraph{FL Implementation.} We use Flower~\citep{flower}, an FL framework that facilitates large-scale experiments through its simulation tools, to implement our setup.
For all the federated models except the FL models with DP-FedAvg, we collaboratively train an LR model with $10,000$ clients for $100$ rounds. At each round of training, we select $10\%$ of the clients randomly to participate in the training. At the end of each round, the parameters are aggregated with the FedAvg algorithm~\citep{mcmahan2017communication}. The optimal number of rounds was determined by following the evolution of the validation loss of different models during training whereas the number of clients to sample for training was chosen with the help of hyperparameters tuning. For the FL model trained with DP-FedAvg, we experiment with
$(50, 100, 200, 1000, 2000)$ clients sampled at each round and  $(25, 50, 75, 100, 125, 150)$ rounds of training since these parameters have an impact on the overall privacy budget.
The best hyperparameters for the federated models have been chosen using a random search.

\paragraph{Warm-up data creation.} To implement the data-sharing strategy needed for training with non-IID data, we split the training set into three while making sure that each set contains different users: 10\% of the dataset is randomly selected to create the warm-up data, and the rest is split between a training set (81\%) and a validation set (9\%). To ensure that no bias came from the warm-up split, we repeat the process three times and test our model with every split. 

\paragraph{Client initialization.} We create each client by randomly selecting one user from the training set. In our dataset, each user corresponds to a unique conversation, either predatory (OG user), and therefore constituted of multiple segments of data, or non-predatory (non-OG user), and therefore constituted of a unique segment of data. If the selected user has a negative label (non-OG user), we select $10$ additional users with negative labels and combine their data to compensate for the lack of non-grooming examples. Note that this was done to compensate for the shortcomings of the PANC dataset and will not be applied in a real-life scenario. 

Finally, at initialization, each client receives a random, balanced portion of the warm-up data: $10$ segments with a negative label and $10$ segments with a positive one to complement their own data. Table~\ref{tab:tabflstat} presents the average number of segments held by each type of client in a federated setting in our experiments.


\begin{table}[t]
\centering
\begin{tabular}{c rr lll rr}
\toprule
      & \multicolumn{2}{l}{Number of segments}     \\
\midrule
Label & OG user           & non-OG user \\
\midrule
 0     & $10$  & $21$ \\
1     & $18$ $(\pm{14})$  & $10$ \\
\bottomrule
\end{tabular}
\caption{Statistics about the data distribution in a federated setting. The training set contains 15,125 non-predatory conversations and 86 predatory conversations.}
\label{tab:tabflstat}
\end{table}

While in our implementation we split the same training data into a ``public'' warm-up set and a ``private'' training set, in a real-life scenario, we expect the warm-up data to come from a publicly available data source 
(for example, a synthetic dataset or an existing publicly available dataset like the one we are using in this paper) since it will be shared among all participants in the training process.

\paragraph{Training with FL with metric DP} 
Since the meaning of the privacy parameters $\eta$ for metric DP depends on the chosen distance measure, the amount of noise added cannot be compared to the traditional $\epsilon$ in DP nor provide the same privacy guarantees. 
To choose the optimal amount of noise $\eta$ for our application, we implement an adversarial attack -- token embedding inversion~\citep{privacyadativebert}. 
We add noise $\eta$ to each embeddings in our training set to obtain perturbed embeddings. For each perturbed embedding, we then compute the closest neighbor among all the non-perturbed embeddings in the training set using the Euclidean distance. We do this for different levels of $\eta$ $(5, 10, 15, 20, 25, 30,$ $35, 40, 45, 50, 55)$. The goal of the attack is to recover the original non-perturbed embedding from the perturbed embedding. 

\begin{figure}[t]
  \centering
  \includegraphics[width=0.8\columnwidth]{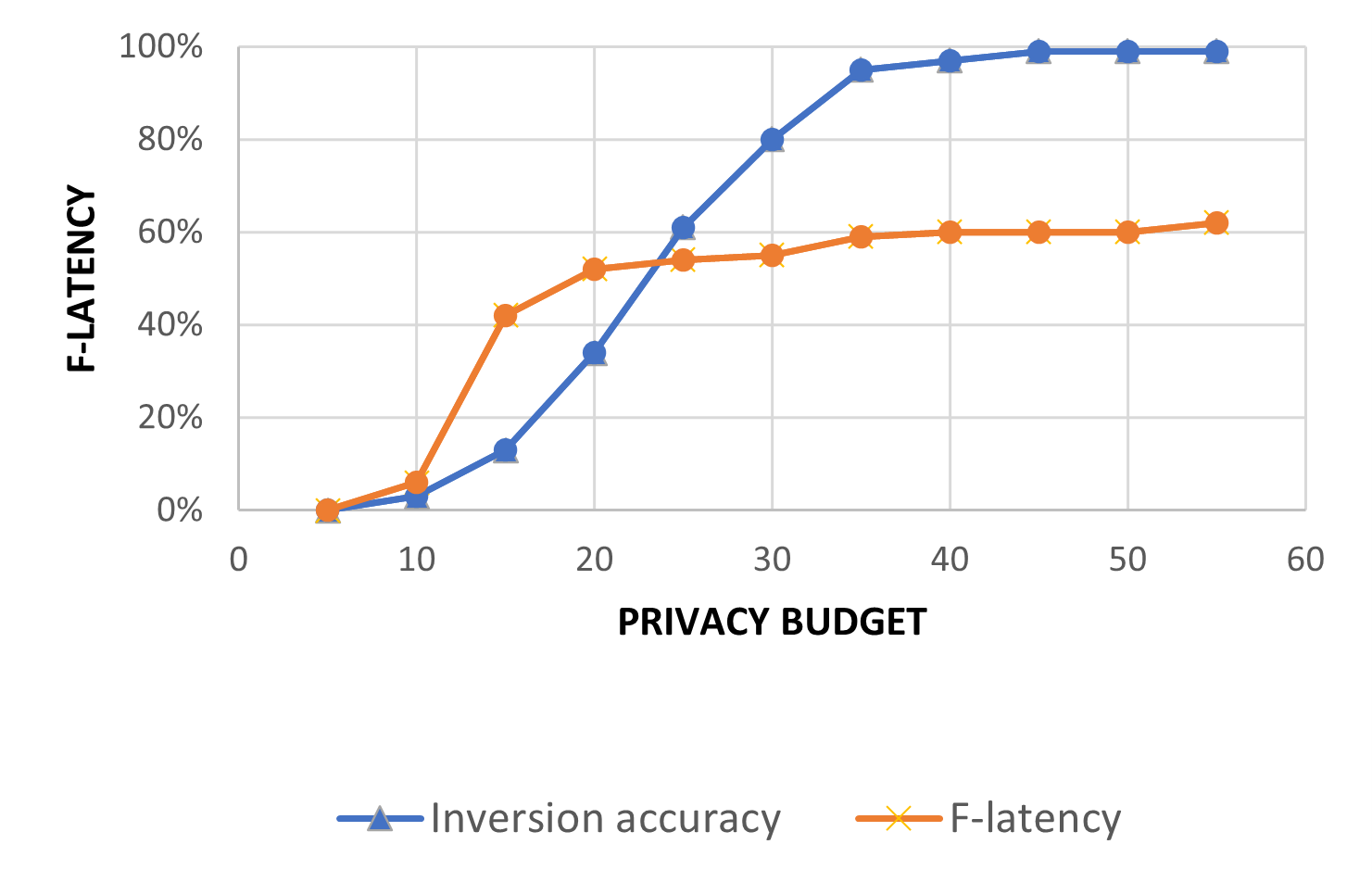}
  \caption{Impact of the privacy budget $\eta$ on the F-latency score of the FL model with metric DP and on the accuracy of the embeddings inversion attack.}
\label{fig:figureldpprivacy}
\end{figure}

Figure~\ref{fig:figureldpprivacy} shows how different levels of noise impact the accuracy of our attack: when $\eta=5$, it is not possible to recover any of the original embeddings
using the perturbed embeddings, whereas when $\eta=55$ the inversion attack is successful at predicting the original embeddings 99\% of the time. Hence, for our application, we estimate that $\eta=20$ offers sufficient protection against a potential attacker since our inversion model is only able to predict the original embeddings with 34\% accuracy.

Furthermore, to alleviate the high cost in utility that comes with adding noise directly to the data, we over-sample the minority class during the training of our models using \citet{errecalde2017temporal}'s oversampling technique. They considered that the minority class is formed not only by the complete conversation but also by portions of the full conversation at different time steps. Therefore, to account for the sequential nature of the eSPD problem and mitigate the imbalanced nature of the data, we enrich our dataset with chunks of conversations from the minority class, in our case, the conversations with a predator. By giving our system more training examples of the beginning of a conversation with a predator, we are able to manage the utility cost of metric-DP. 

\paragraph{Training with FL with DP-SGD} 

We use the Opacus library to compute the privacy budget, with a user-level computation for each round of FL training. Note that the total privacy budget upon training completion may vary if the same user is resampled multiple times. In such scenarios, privacy loss for users participating multiple times in FL training can be computed using composition theorems, such as \cite{kairouz2015composition} and can further amplified by randomized check-ins~\cite{balle2020privacy}. Although we didn't implement a moment accountant on the server to distribute the privacy budget and calculate a tighter bound for composition, our methodology in this study involves using the moment accountant on the client side to manage the privacy budget of DP-SGD. It's important to emphasize that in real-life scenarios, the likelihood of the same user being resampled multiple times is low due to the larger pool of total users available for training and the possibility for each user to opt out of training. 
We conduct a random grid search to select the best hyperparameters: notably, the gradient clipping level $(0.5,1,2,5,7)$, the client's learning rates $(0.01, 0.05, 0.001, 0.0001)$, the batch size $(8, 16, 32, 100)$ and the number of local epochs of training $(1, 5, 10, 15, 20, 100)$.

\paragraph{Training with FL with DP-FedAvg.} We explore DP-FedAvg with both fixed and adaptive clipping~\citep{adaptiveclipping} and test for different values for the update clipping level $(0.1, 0.5, 1, 2, 5, 10)$, noise level $(0.001, 0.01, 0.1, 1, 10,  100)$ and for the number of local epochs of training $(1, 5, 10, 15, 20, 100)$. 


\paragraph{eSPD inference.} All the models were evaluated using a $50$-message sliding window and a skepticism level of $5$, i.e. $5$ of the last $10$ predictions had to be positive before a warning was raised. 

\section{Empirical Results}
\label{sec:addresults}
In this section, we explore additional quantitative results.

\subsection{Warning Latency}
\begin{figure}
    \centering
    \includegraphics[width=0.5\linewidth]{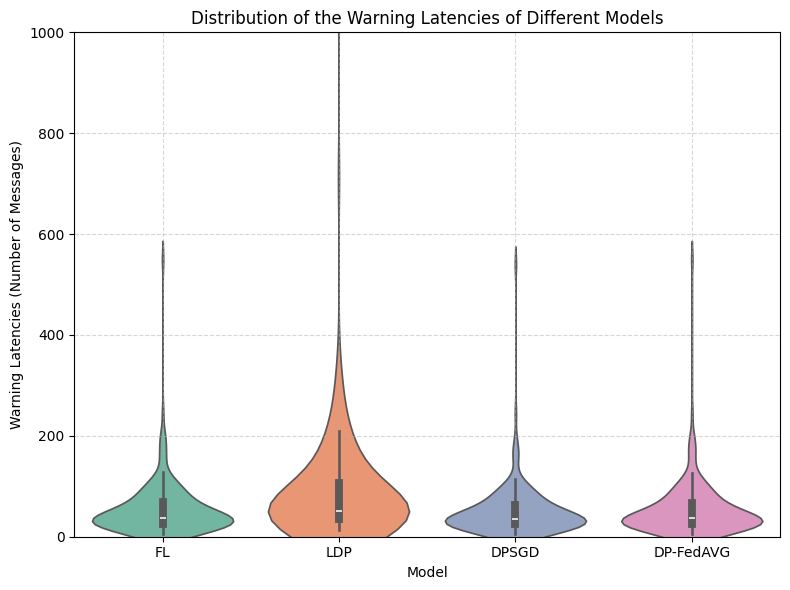}
    \caption{Distribution of the warning latencies for the full predatory conversations for each of our private models.}
    \label{fig:distwarninglat}
\end{figure}

\citet{vogtespd} define the warning latency as the number of messages exchanged before a warning is raised. In Figure~\ref{fig:distwarninglat}, we can see the distribution of the warning latencies for the four private models. We notice that while the FL, DP-SGD and DP-FedAvg models seem to have a similar distribution, for the LDP model, the number of messages it takes to get a final classification is on average higher. At the same time, the maximum of messages reach is also 3 times higher (1531) than the maximum needed for the other models (around 600).

\subsection{Example of Conversation}
Similarly, when looking at the difference in speed between models at the conversation level, we notice differences between the different models. For example, the following predatory conversation is flagged from the get-go by the centralized model after the following messages: 
\begin{itemize}
    \item Predator: ``What’s up little girl?''
    \item Girl: ``Nuttin just chattin''
    \item Predator: ``Kool''
    \item Predator: ``So how old are you''
    \item Girl: ``12/f/nva''
    \item Girl: ``U?''
    
\end{itemize}

While the FL model and the FL model with DPSGD only flag it a dozen messages later, despite exchanges like ``are you not going to talk to me cause i'm older than you'' and only flags it after ``so what you do this past summer''. While the DP-FedAvg model flags it on the first message and the FL model with metric DP only raises a warning much later. This is consistent with the distribution of the warning latencies observed in Figure~\ref{fig:distwarninglat}.

\subsection{False Positive Evaluation}
\begin{figure}
    \centering
    \includegraphics[width=0.5\linewidth]{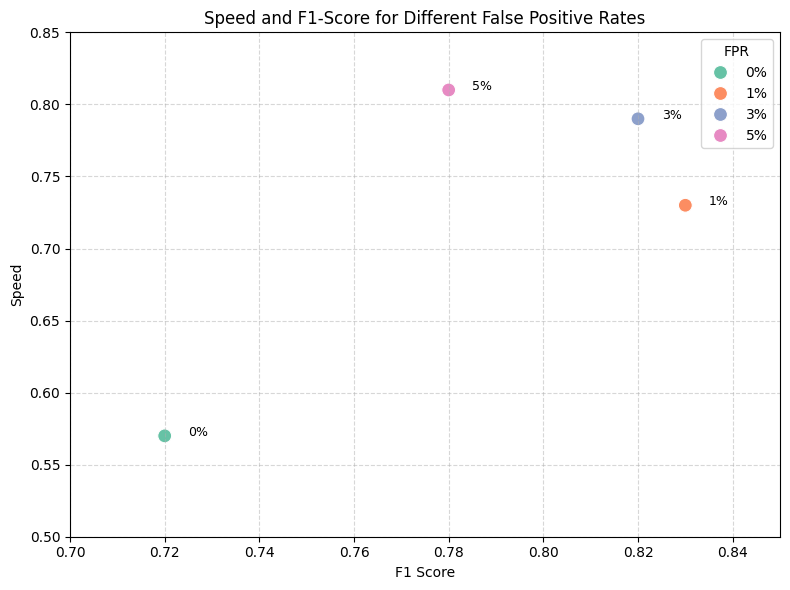}
    \caption{Change in Utility and Speed of Detection for different False Positive Rates}
    \label{fig:fpr}
\end{figure}

In this paper, we stated that a 1\% FPR strikes a good balance between ensuring system efficiency and avoiding unintended discrimination. In Figure~\ref{fig:fpr}, we can see how different false positive rates impact the F1-score and speed of our federated learning model. 

Models with higher FPR are faster but also lose in utility. Depending on the area where the model is deployed and the quality of the classifier, different thresholds might be more appropriate.

\section{Comparison with the literature}
\label{sec:comparisonlit}

While \citet{vogtespd} report their best results with a fine-tuned $BERT_{BASE}$ model, we do not use this model to be able to scale our experiments as fine-tuning large language models with a large number of clients in a federated setting is still an open-research question~\citep{10.1007/978-3-030-80599-9_2}.

However, for comparison purpose, we fine-tune a $BERT_{mobile}$ model in a federated setting and show that we can achieve similar performance with an F-latency score of 58\% than the centralized setting presented by \citet{vogtespd} as seen in Table~\ref{tab:compresults}.

For the rest of our experiments however, we leverage logistic regression to be able to scale to a large number of clients.

\begin{table*}
\centering
{%
\begin{tabular}{lccccc}
\hline
\textbf{Model} & \textbf{F1} & \textbf{Rec} & \textbf{Prec} & \textbf{Speed} & \textbf{F-lat}  \\
\hline
Centralized Fine-Tuned $BERT_{base}$~\citep{vogtespd} & 0.89 & 0.96 & 0.82 & 0.91 & 0.81 \\
Centralized Fine-Tuned $BERT_{mobile}$~\citep{vogtespd} & 0.80 & 0.95 & 0.69 & 0.72 & 0.58 \\
Our FL Fine-Tuned $BERT_{mobile}$  & 0.79 & 0.96 & 0.68 & 0.73 & 0.58 \\ 

\hline
\end{tabular}}
\caption{Evaluation results for the eSPD task - Baselines and Comparison with the literature}
\label{tab:compresults}

\end{table*}

\end{document}